\documentclass{article}
\usepackage{ijcai23}

\usepackage{times}
\usepackage{soul}
\usepackage{url}
\usepackage[hidelinks]{hyperref}
\usepackage[utf8]{inputenc}
\usepackage[small]{caption}
\usepackage{graphicx}
\usepackage{amsmath}
\usepackage{amsthm}
\usepackage{booktabs}
\usepackage{algorithm}
\usepackage{algorithmic}
\usepackage[switch]{lineno}
\usepackage{subfig}
\usepackage{multirow}
\usepackage{amsmath}
\usepackage{amsfonts} 
\usepackage{bbding}
\usepackage{makecell}
\usepackage{colortbl}
\usepackage[table]{xcolor}
\usepackage{microtype}
\usepackage{bm}
\usepackage{booktabs}
\usepackage{wrapfig}
\usepackage{subfiles}
\definecolor{cred}{HTML}{FF6B6B}
\definecolor{cyellow}{HTML}{FEC260}
\definecolor{cgreen}{HTML}{70AD47}
\definecolor{cblue}{HTML}{4D96FF}
\definecolor{cpurple}{HTML}{2A0944}
\definecolor{ggray}{RGB}{127,127,127}
\definecolor{aliceblue}{rgb}{0.94, 0.97, 1.0}

\newcommand{\myparagraph}[1]{\textbf{#1}\hspace{1.8ex}}

\title{Text-Video Retrieval with Disentangled Conceptualization and Set-to-Set \\Alignment}

\author{
    Peng Jin$^{1,2}$\and
    Hao Li$^{1,2}$\and
    Zesen Cheng$^{1,2}$\and
    Jinfa Huang$^{1,2}$\and 
    Zhennan Wang$^{3}$\and \\
    Li Yuan$^{1,2,3}$\and
    Chang Liu$^{4}$\footnotemark[2]\and
    Jie Chen$^{1,2,3}$\footnotemark[2] 
    \affiliations
    \small{$^1$School of Electronic and Computer Engineering, Peking University, Shenzhen, China} \\
    \small{$^2$AI for Science (AI4S)-Preferred Program, Peking University Shenzhen Graduate School, Shenzhen, China} \\
    \small{$^3$Peng Cheng Laboratory, Shenzhen, China} \quad \small{$^4$Department of Automation and BNRist, Tsinghua University, Beijing, China}
    \emails
    \small{\{jp21,  cyanlaser, jinfahuang\}@stu.pku.edu.cn \quad \{lihao1984, yuanli-ece\}@pku.edu.cn}\\ \small{wangzhennan2017@email.szu.edu.cn \quad liuchang2022@tsinghua.edu.cn \quad chenj@pcl.ac.cn}
}

\makeatletter
\newcommand{\ssymbol}[1]{$^{\@fnsymbol{#1}}$}
\makeatother

\begin{document}

\maketitle
\begin{abstract}
    Text-video retrieval is a challenging cross-modal task, which aims to align visual entities with natural language descriptions. Current methods either fail to leverage the local details or are computationally expensive. What's worse, they fail to leverage the heterogeneous concepts in data. In this paper, we propose the Disentangled Conceptualization and Set-to-set Alignment (DiCoSA) to simulate the conceptualizing and reasoning process of human beings. For disentangled conceptualization, we divide the coarse feature into multiple latent factors related to semantic concepts. For set-to-set alignment, where a set of visual concepts correspond to a set of textual concepts, we propose an adaptive pooling method to aggregate semantic concepts to address the partial matching. In particular, since we encode concepts independently in only a few dimensions, DiCoSA is superior at efficiency and granularity, ensuring fine-grained interactions using a similar computational complexity as coarse-grained alignment. Extensive experiments on five datasets, including MSR-VTT, LSMDC, MSVD, ActivityNet, and DiDeMo, demonstrate that our method outperforms the existing state-of-the-art methods.
\end{abstract}
\footnotetext[2]{Corresponding author: Chang Liu, Jie Chen.}
\footnotetext[1]{ Code is available at \href{https://github.com/jpthu17/DiCoSA}{https://github.com/jpthu17/DiCoSA}.}

\section{Introduction}
\begin{figure}[t]
    \centering
    %\vspace{-0.5em}
    \includegraphics[width=1.\linewidth]{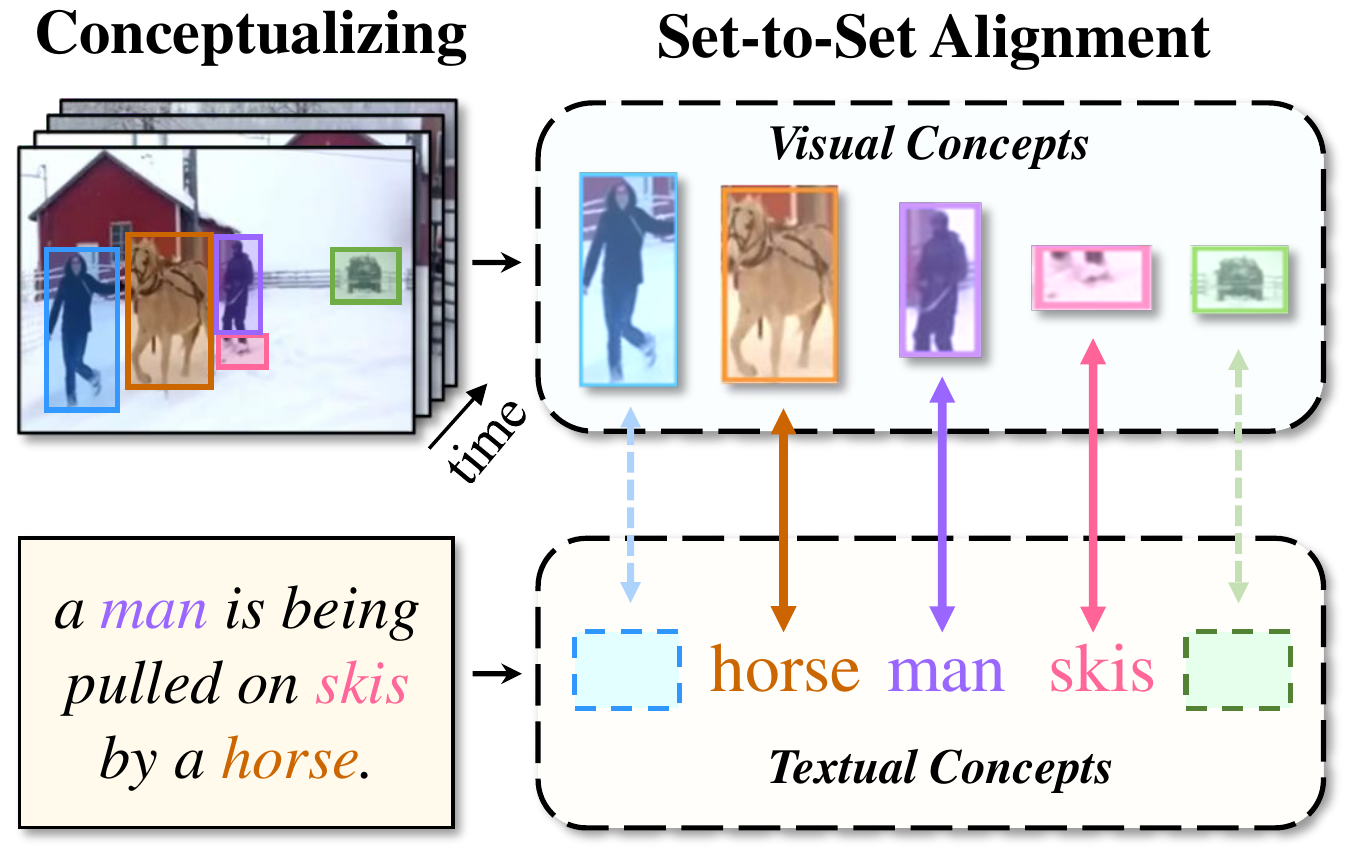}
    \vspace{-0.8em}
    \caption{In text-video retrieval, a set of visual concepts corresponds to a set of textual concepts. More importantly, the semantic concepts across text and video are typically partially matched. For example, the description is not exhaustive and does not describe the visual concepts, i.e., ``woman'' and ``car''.}
    \vspace{-.5em}
    \label{fig1}
\end{figure}

Recent years have witnessed encouraging progress in text-video retrieval, which enables humans to associate textual concepts with video entities and vice versa~\cite{wang2021t2vlad,jin2023diffusionret}. As illustrated in Figure~{\color{red}{\ref{fig1}}}, humans perceive the cross-modal matching task by conceptualizing high-dimensional inputs from multiple modalities and reasoning with concepts to achieve partially matched set-to-set alignment. In stark contrast, machine models typically represent each modality as a perceptual whole.

\begin{figure*}[thb]
    \centering
    \includegraphics[width=1.0\linewidth]{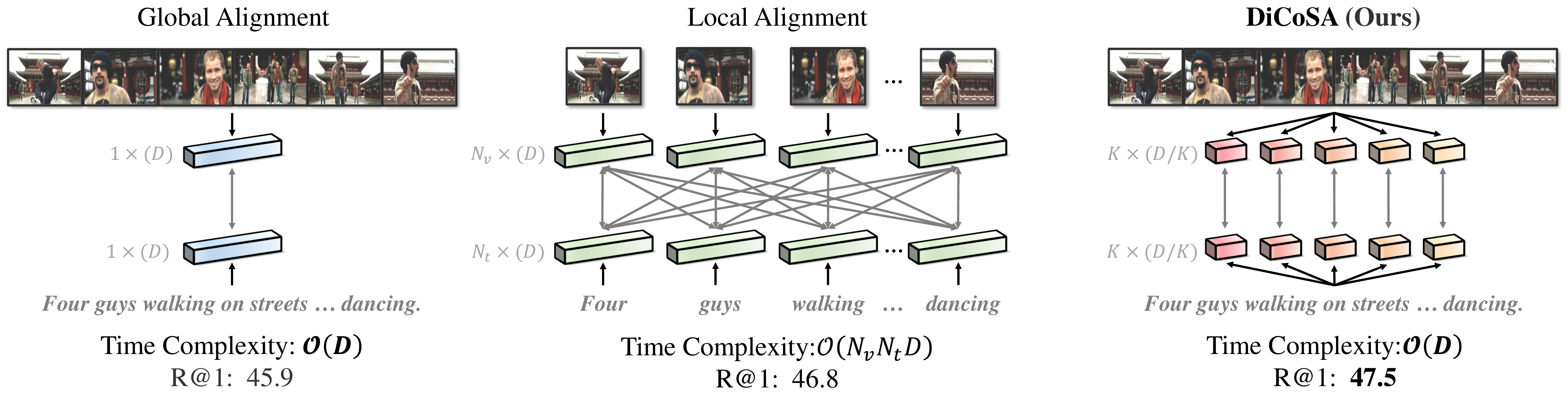}  
    \vspace{-1.5em}
    \caption{\textbf{Global alignment} (\textcolor{cblue}{\textbf{left}}) treats discriminative regions equally and fails to capture local details. \textbf{Local alignment} (\textcolor{cgreen}{\textbf{middle}}) is computationally expensive due to the exhaustive matching operation. The previous methods typically characterize semantic concepts of texts and videos in a state of entanglement. In stark contrast, our \textbf{DiCoSA} (\textcolor{cred}{\textbf{right}}) aligns texts and videos on heterogeneous semantic concepts to achieve humanlike set-to-set matching. Specifically, we divide the coarse feature into multiple latent factors related to semantic concepts~(e.g., ``man'' and ``skis''). Further, we propose adaptive pooling to aggregate semantic concepts to address the partial matching in set-to-set alignment. Since we encode concepts independently in only a few dimensions ($\mathbb{R}^{\frac{D}{K}}$), our method is superior at computation efficiency and granularity. In addition, We also analyze time complexity and text-to-video retrieval performance at R@1 above in this figure. Here, $D$, $N_v$, and $N_t$ denote the feature dimension, the frame length, and the text length, respectively.}
    \label{fig2}
    \vspace{-0.5em}
\end{figure*}

Existing methods for text-video retrieval mainly focus on learning a joint feature representation space for different modalities, where text-video similarities could be measured to enable cross-modal matching. Such cross-modal representation learning methods allow for both global alignment~\cite{liu2019use,gabeur2020multi} and local alignment~\cite{wang2022disentangled,wray2019fine,chen2020fine} between text and video feature representations. The global alignment methods exploit high-level semantics for text-video retrieval. For example, CLIP4Clip~\cite{luo2021clip4clip} adopts the text-image pre-training model CLIP~\cite{radford2021learning} to transfer the knowledge for enhancing global representation. To leverage the local details, the local alignment methods study fine-grained semantic alignment for text-video retrieval. For instance, T2VLAD~\cite{wang2021t2vlad} shows the potential of local alignment which aligns each word and each frame individually to improve fine-grained retrieval.

However, existing methods are deficient in the following three aspects. (i) The \textbf{global alignment} (Figure~{\color{red}\ref{fig2}}, left) may treat discriminative regions equally and fail to capture local details in texts and videos. (ii) The \textbf{local alignment} (Figure~{\color{red}\ref{fig2}}, middle) aligns each word and per frame individually and is computationally expensive due to the exhaustive matching operation. (iii) \textbf{Both global and local alignments} coarsely represent the text (video) as a perceptual whole encoded by a set of concepts. As a result, they may fail to leverage the heterogeneous concepts in data and tend to focus on identifying the invariant features. As an example to illustrate, the video in Figure~{\color{red}\ref{fig1}} involves a set of concepts such as ``man'', ``skis'', ``woman'', and ``horse''. To align with the text query, the video representations learned by the existing coarse approaches may continuously rely on the salient factors ``man'' and ``horse'', and yet ignore other important factors (e.g., ``woman'' and ``skis'') and the relations among these factors. This flaw might make these approaches stuck in the local invariant matching and misunderstand the cross-modal interaction and context.

A reasonable solution to tackle the cross-modal matching task is to align texts and videos on heterogeneous semantic concepts~(Figure~{\color{red}\ref{fig2}}, right). The core insight is simulating the human process of conceptualizing things and reasoning on the sets of concepts. To learn explanatory and discriminative factors of variations, we adopt disentangled representations learning. A disentangled representation independently encodes information about each latent factor in only a few dimensions.

To this end, we propose the \underline{Di}sentangled \underline{Co}nceptualization and \underline{S}et-to-set \underline{A}lignment~(DiCoSA), as shown in Figure~{\color{red}\ref{fig3}}. In detail, we disentangle high-dimensional coarse features into compact latent factors which explicitly encode textual semantics and visual entities. Then, we optimize latent factors from both inter-concept and intra-concept perspectives. In the inter-concept perspective, we minimize the inter-concept mutual information to find representation subspaces with minimal relevance to each other for decoupling representation. In the intra-concept perspective, we maximize the mutual information of each latent factor pair separately to align language and video within each concept. However, due to the information across modalities typically being only partially matched~\cite{liu2021adaptive}, we cannot blindly leverage superficial correlations between latent factors for text-video retrieval. To address this problem, as shown in the bottom panel of Figure~{\color{red}\ref{fig3}}, we propose an uncertainty-aware module to estimate the confidence of each cross-modal concept matching. Finally, we use the confidence as the weight to aggregate all factor pairs to calculate the similarity of text and video, which is called adaptive pooling.  

In particular, since the dimension of the latent factor ($\mathbb{R}^{\frac{D}{K}}$) is lower than the original feature dimension ($\mathbb{R}^{D}$), our method ensures fine-grained interactions with local alignment using a similar computational complexity as global alignment. Our contributions are summarized as follows: 
\begin{figure*}[thb]
    \centering
    \includegraphics[width=1.0\linewidth]{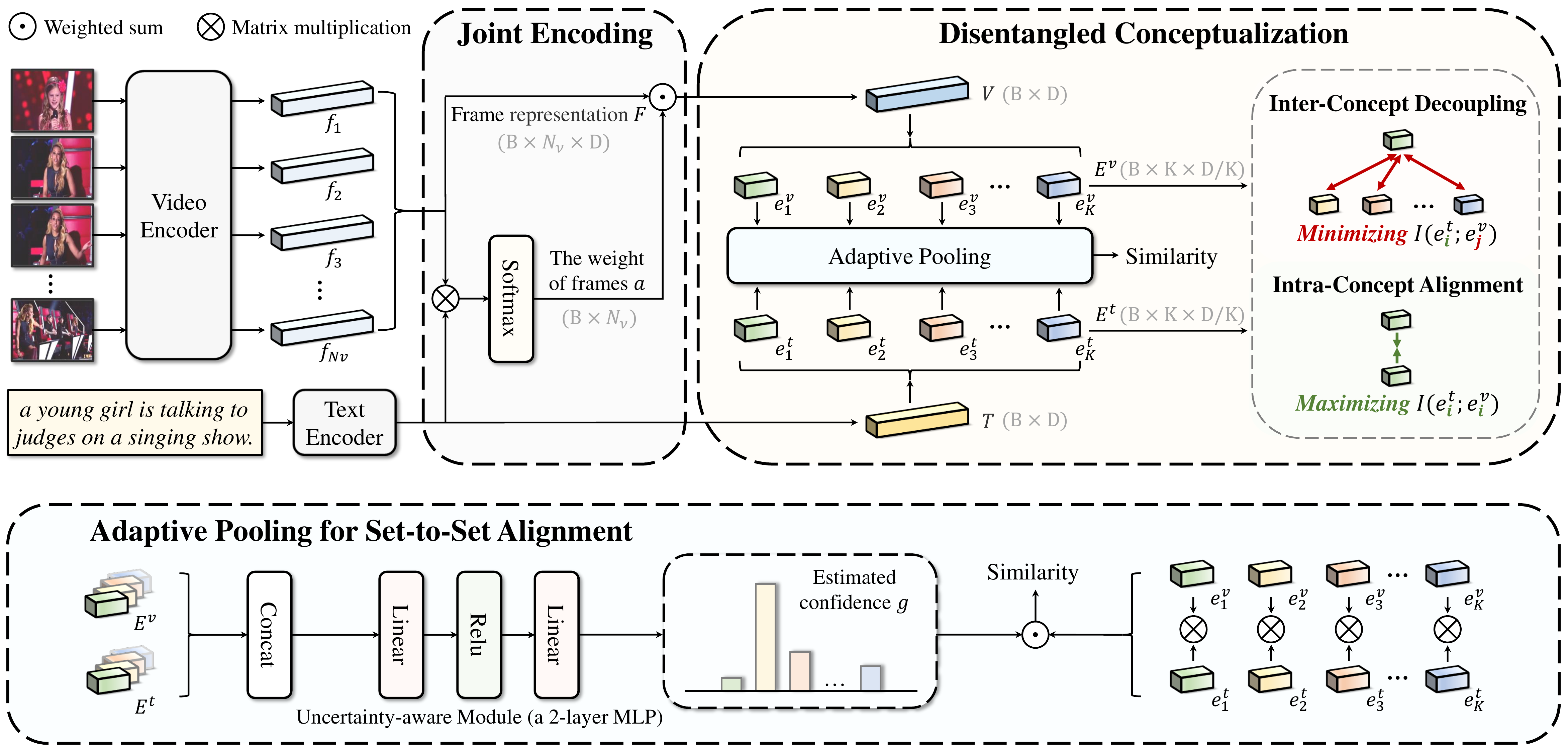}  
    \vspace{-.5em}
    \caption{\textbf{Our DiCoSA framework for text-video retrieval.} Given a text-video pair, we leverage a transformer-based text encoder to extract the global text representation $T$. Likewise, we also use transformer to extract frame representation $F$ and then aggregate frame representation with text as the condition to obtain video representation $V$. Further, we separate the latent factors $E^v$ of the video representation and $E^t$ of the text representation. To capture disentangled representation, we optimize latent factors from both inter-concept and intra-concept perspectives. Due to the information across modalities being partially matched, we aggregate semantic concepts via adaptive pooling to achieve set-to-set alignment, as shown in the bottom panel of this figure. Here, $B$, $D$, $K$, $N_v$ and $N_t$ denote the batch size, the original feature dimension, the number of concepts, the frame length and the text length, respectively.}
    \label{fig3}
    \vspace{-0.5em}
\end{figure*}
\begin{itemize}

\item We propose a disentangled conceptualization method for text-video retrieval, which divides the coarse features into multiple latent factors related to semantic concepts and achieves humanlike set-to-set matching. 
\item To address the partial matching of visual entities and various phrases, we propose adaptive pooling to locate mismatched cross-modal concepts and aggregate all factor pairs to calculate the similarity of text and video.
\item We conduct extensive experiments on five datasets, i.e., MSR-VTT, LSMDC, MSVD, ActivityNet and DiDeMo, and achieve new state-of-the-art retrieval performance.
\end{itemize}

\section{Related Work}
\myparagraph{Text-Video Retrieval.}
Most works~\cite{jin2022expectation,jin2023diffusionret} of text-video retrieval is based on contrastive learning~\cite{zhang2021zero,zhang2022m,zhang2022align,zhang2023patch} to map text and video to the same semantic space. For example, CLIP-ViP~\cite{xue2022clip} explores factors that hinder video post-pretraining on pre-trained image-text models and effectively leverage image-text pre-trained model for post-pretraining. HBI~\cite{jin2023video} designs a new framework of multivariate interaction for cross-modal representation learning~\cite{li2022toward,li2022joint,cheng2023parallel,ye2023fits}. DRL~\cite{wang2022disentangled} aligns each word and each frame individually to achieve token-wise interaction. However, a visual entity may relate to multiple frames, and also a frame may relate to multiple visual entities. Therefore, the previous methods cannot achieve element-level matching between semantic concepts. By contrast, our model benefits from the disentangled latent factors for element-level cross-modal matching.

\noindent \myparagraph{Disentangled Representation.}
This concept was first proposed by~\cite{bengio2009learning}, which aims to separate the latent factors of variations behind the data. Recently, disentangled representation learning has received lots of attention~\cite{sreekar2021mutual,wang2021self,tran2017disentangled,suter2019robustly,van2019disentangled,locatello2019disentangling}. However, how to use the disentangled latent factors for fine-grained retrieval remains largely unexplored for text-video retrieval. Based on the information-theoretic view~\cite{chen2016infogan,do2019theory}, several works provide a more detailed description~\cite{do2019theory,eastwood2018framework,ridgeway2018learning} by explicitly measuring the properties of a disentangled representation~\cite{eastwood2018framework,ridgeway2018learning}. We adopt the information-theoretic definition and show that the proposed DSCA converges with semantic disentanglement.

\section{Methodology}
We focus on the tasks of text-to-video and video-to-text retrieval. In the text-to-video retrieval task, given a query text $\textbf{t}$ and candidate videos $\textbf{v}$, our goal is to rank all videos according to semantic similarity. Similarly, the goal of video-to-text retrieval is to rank all candidate text based on the query video. The problem is formulated as a modality similarity measurement, where the similarity of matched text-video pairs is as high as possible and the similarity of unmatched pairs is as low as possible. Figure~{\color{red}\ref{fig3}} illustrates the overall framework.

\subsection{Text-Video Joint Encoding}
For text representation, we adopt the text encoder of CLIP (ViT-B/32)~\cite{radford2021learning}. The output from the [CLS] token is taken as the text representation. For the input text $\textbf{t}$, we denote the generated representation as $T\in \mathbb{R}^D$. 

\newcommand{\pub}[1]{\color{gray}{\tiny{#1}}}
\newcommand{\Frst}[1]{{\textbf{#1}}}
\newcommand{\Scnd}[1]{{\underline{#1}}}
\begin{table*}[htb]
\centering
\footnotesize
{
{
\begin{tabular}{p{100pt}|ccccc|ccccc}
\toprule[1.5pt]
\multirow{2}{*}{\textbf{Method}} &  \multicolumn{5}{c|}{\textbf{Text-\textgreater{}Video}}        & \multicolumn{5}{c}{\textbf{Video-\textgreater{}Text}}  \\ 
\cmidrule(rl){2-6}\cmidrule(rl){7-11}
  & {R@1$\uparrow$} & R@5$\uparrow$ & R@10$\uparrow$ & MdR$\downarrow$ & MnR$\downarrow$
& R@1$\uparrow$ & R@5$\uparrow$ & R@10$\uparrow$ & MdR$\downarrow$ & MnR$\downarrow$    \\ \midrule
CE~\pub{\cite{liu2019use}} & {20.9} & 48.8 & 62.4 & 6.0 & 28.2 & 20.6 & 50.3 & 64.0 & 5.3 & 25.1\\
MMT~\pub{\cite{gabeur2020multi}} & {26.6} & 57.1 & 69.6 & 4.0 & 24.0 & 27.0 & 57.5 & 69.7 & 3.7 & 21.3 \\
Support-Set~\pub{\cite{patrick2021support}}  & {30.1} & 58.5 & 69.3 & 3.0 & - &28.5 &58.6 &71.6 &3.0 & - \\
T2VLAD~\pub{\cite{wang2021t2vlad}} & {29.5} & 59.0 & 70.1 & 4.0 & -  & 31.8 &60.0 &71.1 &3.0 & -\\
TT-CE~\pub{\cite{croitoru2021teachtext}} & {29.6} & 61.6 & 74.2 & 3.0 & - & 32.1 & 62.7 & 75.0 & 3.0 & -\\
FROZEN~\pub{\cite{bain2021frozen}} & {31.0} & 59.5 & 70.5 & 3.0 & - & - &- &- &- & -\\
CLIP4Clip~\pub{\cite{luo2021clip4clip}}  & {44.5} & 71.4 & 81.6 &\Frst{2.0} & 15.3 & 42.7 & 70.9& 80.6 & \Frst{2.0} & 11.6\\
CLIP2Video~\pub{\cite{fang2021clip2video}}  & {45.6} & 72.6 & 81.7 &\Frst{2.0} & 14.6 & 43.5 & 72.3& 82.1 & \Frst{2.0} &10.2 \\
EMCL-Net~\pub{\cite{jin2022expectation}}  & {46.8} & 73.1 & 83.1 &\Frst{2.0} & - & 
46.5 & 73.5 & 83.5 & \Frst{2.0} & - \\
X-Pool~\pub{\cite{gorti2022x}}  & {46.9} & 72.8 & 82.2 &\Frst{2.0} & 14.3 & 44.4 & 73.3 & 84.0 & \Frst{2.0} & 9.0 \\
TS2-Net~\pub{\cite{liu2022ts2}}  & {47.0} & 74.5 & \Frst{83.8} & \Frst{2.0} & \Frst{13.0} & 45.3 & 74.1 & 83.7 & \Frst{2.0} & 9.2 \\ 
\midrule
 \rowcolor{aliceblue!60} \textbf{DiCoSA (Ours)}  & {\Frst{47.5}} & \Frst{74.7} & \Frst{83.8} & \Frst{2.0} & 13.2 & \Frst{46.7} & \Frst{75.2} & \Frst{84.3} & \Frst{2.0} & \Frst{8.9}      \\ \bottomrule[1.5pt]
\end{tabular}
}
}
\vspace{-0.5em}
\caption{\textbf{Retrieval performance on the MSR-VTT dataset.} ``$\uparrow$'' denotes that higher is better. ``$\downarrow$'' denotes that lower is better.}
\label{MSRVTT results}
\vspace{-0.5em}
\end{table*}

For video representation, we first evenly extract the frames from the video clip as the input sequence of video $\textbf{v}=\{v_{1},v_{2},...,v_{|N_v|}\}$, where $N_v$ denotes the frame length. Subsequently, we use ViT~\cite{dosovitskiy2021an} to encode the frame sequence. Following CLIP, we adopt the output from the [CLS] token as the frame embedding. After that, we use a temporal transformer (a 4-layer transformer) to aggregate the embedding of all frames and obtain the frame representation $F$. Inspired by~\cite{gorti2022x,bain2022clip}, we aggregate frame representation with text as the condition. In detail, we calculate the inner product between the text representation $T$ and frame representation $F=\{f_{1},f_{2},...,f_{N_v}\}$. We get the weight of the frames by:
\begin{equation}
a_i=\frac{\exp((T)^\top f_i/\tau)}{\sum_{i=1}^{N_v}\exp((T)^\top f_k/\tau)},
\end{equation}
where $\tau$ is the trade-off hyper-parameter. The smaller $\tau$ allows visual features to take more textual conditions into account during aggregation. The final video representation $V\in \mathbb{R}^D$ is defined as $V=\sum_{i=1}^{N_v}a_i f_i$.

\subsection{Disentangled Conceptualization from both Inter-Concept and Intra-Concept Perspectives}
The above encoding methods only generate the holistic text representations $T$ and video representations $V$. These representations characterize semantic concepts of the input texts and videos in a state of entanglement. Therefore, a direct similarity matching of these representations cannot ensure adequate set-to-set cross-modal matching between a set of various phrases contained in the text description and a set of varying visual entities in the video sequence. In the following, we learn explanatory disentangled factors of variations in text and video representations to explicitly measure and understand the cross-modal relevance.

We start with the text representation $T\in \mathbb{R}^D$. Following the setting of disentangled representation learning~\cite{ma2019disentangled}, we assume that each text representation is disentangled into $K$ independent latent factors, i.e., $E^{t}=[e_1^t,e_2^t,...,e_K^t]$. Each latent factor $e_k^t\in \mathbb{R}^{\frac{D}{K}}$ represents a specific semantic concept in the text, and the independence of the latent factors ensures that those semantic concepts are not related to each other. Specifically, we independently project the text representation $T$ into $K$ components, and obtain the $k_{\textrm{th}}$ latent factor $e_k^t$ as follows,
\begin{equation}
e_{k}^t = W_{k}^{t}T,
\end{equation}
where $W_{k}^t\in \mathbb{R}^{\frac{D}{K} \times D}$ is trainable parameter. The latent factor $e_{n}^v$ of video representation can be calculated in the same way, i.e., $e_{k}^v = W_{k}^{v}V$. In this way, we project features explicitly into representation subspaces corresponding to different concepts. The model is then able to optimize and reason information separately from different representation subspaces.

\subsubsection{Inter-Concept Decoupling}
In order to find representation subspaces with minimal relevance to each other and hence improve the respective discriminative power for semantic matching task, we propose to minimize the inter-concept mutual information.  Given two latent factors $e_i^t$ and $e_j^v$, their mutual information is defined in terms of their probabilistic density functions:
\begin{equation}
I(e_i^t;e_j^v)=\mathbb{E}_{\bm{t},\bm{v}} \left[ p(e_i^t,e_j^v)\log\frac{p(e_i^t,e_j^v)}{p(e_i^t)p(e_j^v)}\right].
\end{equation}
Obviously, the mutual information is hard to measure directly. To this end, we implicitly measure the mutual information via an encoder discriminator architecture. Concretely, given latent factors $e^t\in \mathbb{R}^{\frac{D}{K}}$ and $e^v\in \mathbb{R}^{\frac{D}{K}}$, we first normalize them by the following formula:
\begin{equation}
z^t=\frac{e^t-\mathbb{E}\left[e^t\right]}{\sqrt{\textrm{Var}\left[e^t\right] }},\quad z^v=\frac{e^v-\mathbb{E}\left[e^v\right]}{\sqrt{\textrm{Var}\left[e^v \right] }},
\end{equation}
where $z^t, z^v$ have the same mean and standard deviation. In this manner, we scale the latent factors to the standard scale. Then, we calculate the covariance of $z^t$ and $z^v$ as follows,
\begin{equation}
C_{i,j}=\mathbb{E}_{\bm{t},\bm{v}} \left[ (z^t_i)^{\top}z^v_j \right],
\end{equation}
where $z^t_i$ and $z^v_j$ are the normalized features of $e^t_i$ and $e^v_j$, respectively.

\newtheorem{myTheo}{Lemma}
\begin{myTheo}\label{thm1}
  Maximizing (minimizing) $I(e_i^t;e_j^v)$ is equivalent to maximizing (minimizing) $C_{i,j}$, i.e., $ C_{i,j}\propto I(e_i^t;e_j^v)$. 
\end{myTheo}

We refer the reader to our supplemental material for more detail about Lemma 1. The final inter-concept decoupling loss $\mathcal{L}_{D}$ is calculated as follows,
\begin{equation}
\mathcal{L}_{D}=\sum_i \sum_{j\neq i} (C_{i,j})^2.
\end{equation}
This loss minimizes the mutual information of negative pairs $(e_i^t,e_j^v)$, thereby decoupling the latent factors.

\subsubsection{Intra-Concept Alignment}
To comprehensively capture $K$ latent factors from both text and video representations, we are required to mine semantic concepts from the text-video pairs. To this end, we optimize the representation subspace corresponding to each latent factor separately. The key insight here is that we consider each representation subspace independently, instead of text-video pairs, to comprehensively describe the latent factors and capture their relevance. Specifically, we maximize the mutual information between the text latent factor and the corresponding video latent factor within the same subspaces. The intra-concept alignment loss $\mathcal{L}_{A}$ is formulated as:
\begin{equation}
\mathcal{L}_{A}=\sum_i (1-C_{i,i})^2.
\end{equation}
This loss maximizes the mutual information of each positive pair $(e_i^t,e_i^v)$ separately. 

\begin{table*}[htb]
\centering
\footnotesize
%\resizebox{1.\linewidth}{!}
{
\begin{tabular}{p{100pt}|ccccc|ccccc}
\toprule[1.5pt]
\multirow{2}{*}{\textbf{Method}} &  \multicolumn{5}{c|}{\textbf{LSMDC}}        & \multicolumn{5}{c}{\textbf{MSVD}}  \\ 
\cmidrule(rl){2-6}\cmidrule(rl){7-11}
  & R@1$\uparrow$ & R@5$\uparrow$ & R@10$\uparrow$ & MdR$\downarrow$ & MnR$\downarrow$
& R@1$\uparrow$ & R@5$\uparrow$ & R@10$\uparrow$ & MdR$\downarrow$ & MnR$\downarrow$    \\ \midrule
FROZEN~\pub{\cite{bain2021frozen}} & 15.0 & 30.8 & 39.8 & 20.0 & - & 33.7 & 64.7 & 76.3 & 3.0 & -\\
CLIP4Clip~\pub{\cite{luo2021clip4clip}}  & 22.6 & 41.0 & 49.1 & 11.0 & 61.0 & 45.2 & 75.5 & 84.3 & \Frst{2.0} & 10.3 \\
EMCL-Net~\pub{\cite{jin2022expectation}}  & 23.9 & 42.4 & 50.9 & 10.0 & - & - & - & - & - & - \\
TS2-Net~\pub{\cite{liu2022ts2}}  & 23.4 & 42.3 & 50.9 & 9.0 & 56.9 & - & - & - & - & - \\
X-Pool~\pub{\cite{gorti2022x}}  & 25.2 & \Frst{43.7} & 53.5 & \Frst{8.0} & 53.2 & 47.2 & \Frst{77.4} & \Frst{86.0} & \Frst{2.0} & 9.3 \\
\midrule
\rowcolor{aliceblue!60} \textbf{DiCoSA (Ours)}  & \Frst{25.4} & 43.6 & \Frst{54.0} & \Frst{8.0} & \Frst{41.9} & \Frst{47.4} & 76.8 & \Frst{86.0} & \Frst{2.0} & \Frst{9.1}      \\  %\bottomrule
\midrule[1.25pt]
\multirow{2}{*}{\textbf{Method}} &  \multicolumn{5}{c|}{\textbf{ActivityNet}}        & \multicolumn{5}{c}{\textbf{DiDeMo}}  \\ 
\cmidrule(rl){2-6}\cmidrule(rl){7-11}
  & R@1$\uparrow$ & R@5$\uparrow$ & R@10$\uparrow$ & MdR$\downarrow$ & MnR$\downarrow$
& R@1$\uparrow$ & R@5$\uparrow$ & R@10$\uparrow$ & MdR$\downarrow$ & MnR$\downarrow$    \\ \midrule
CE~\pub{\cite{liu2019use}} & 18.2 & 47.7 & 63.9 & 6.0 & 23.1 & 16.1 & 41.1 & - & 8.3 & 43.7\\
ClipBERT~\pub{\cite{lei2021less}} & 21.3 & 49.0 & 63.5 & 6.0 & - & 20.4 & 48.0 & 60.8 & 6.0 & - \\
TT-CE~\pub{\cite{croitoru2021teachtext}}  & 23.5 & 57.2 & - & 4.0 & - & 21.6 & 48.6 & 62.9 & 6.0 & - \\
CLIP4Clip~\pub{\cite{luo2021clip4clip}}  & 40.5 & 72.4 & 83.6 & \Frst{2.0} & 7.5 & 42.8 & 68.5 & 79.2 & \Frst{2.0} & 18.9 \\
TS2-Net~\pub{\cite{liu2022ts2}}  & 41.0 & \Frst{73.6} & 84.5 & \Frst{2.0} & 8.4 & 41.8 & 71.6 & 82.0 & \Frst{2.0} & 14.8 \\
\midrule
\rowcolor{aliceblue!60} \textbf{DiCoSA (Ours)}  & \Frst{42.1} & \Frst{73.6} & \Frst{84.6} & \Frst{2.0} & \Frst{6.8} & \Frst{45.7} & \Frst{74.6} &\Frst{83.5} &\Frst{2.0} &\Frst{11.7} \\ \bottomrule[1.5pt]
\end{tabular}
}
\vspace{-0.5em}
\caption{\textbf{Text-to-video retrieval performance on other datasets.} ``$\uparrow$'' denotes that higher is better. ``$\downarrow$'' denotes that lower is better.}
\label{Experiments0}
\vspace{-0.5em}
\end{table*}

\subsection{Set-to-Set Alignment via Adaptive Pooling}
However, the information across text and video is typically partially matched~\cite{liu2021adaptive}. We hence cannot directly and blindly leverage superficial correlations between latent factors for set-to-set alignment. To this end, we propose adaptive pooling to locate those mismatched cross-modal concepts and reduce their impact on the final similarity calculation.

To reveal the mismatched cross-modal latent factors, we design an uncertainty-aware module to estimate the confidence of each cross-modal concept matching. Specifically, we concatenate latent factor $e_i^t\in \mathbb{R}^{\frac{D}{K}}$ with the latent factor $e_i^v\in \mathbb{R}^{\frac{D}{K}}$ in the $i_{th}$ subspace, generating the input data of the uncertainty-aware module $\hat e_i=[e_i^t, e_i^v]\in \mathbb{R}^{\frac{2D}{K}}$. Then, the confidence of the $i_{th}$ subspace is obtained by:
\begin{equation}
g_i=\textrm{MLP}(\hat e_i),
\end{equation}
where ``MLP'' consists of two linear layers and an activation function. Usually, small $g_i$ indicates that the concept corresponding to the $i_{th}$ subspace is matched with a low probability. Then, we use the confidence as the weight to aggregate all factor pairs to calculate the similarity of text and video, which is called adaptive pooling. Finally, the similarity is defined as:
\begin{equation}
S=\sum_{i=1}^K g_i \frac{{(e_i^t)}^\top e_i^v}{\left\|e^t_i\right\|\left\|e^v_i\right\|}.
\end{equation}

\subsection{Training Objective}
Following common practice, we leverage InfoNCE loss~\cite{oord2018representation} to optimize cross-modal similarity:
\begin{equation}
\begin{aligned} 
\mathcal{L}_{S}=-\frac{1}{2}(\frac{1}{B}\sum_{l=1}^{B}\log\frac{\exp(S_{l,l}/\tau^{'})}{\sum_{k=1}^{B}\exp(S_{l,k}/\tau^{'})}+\\
\frac{1}{B}\sum_{k=1}^{B}\log\frac{\exp(S_{k,k}/\tau^{'})}{\sum_{l=1}^{B}\exp(S_{l,k}/\tau^{'})}),
\end{aligned}
\end{equation}
where $B$ is the batch size and $\tau^{'}$ is a pre-defined temperature prior. $S_{l,k}$ is the similarity between the $l_{\textrm{th}}$ text and the $k_{\textrm{th}}$ video. Combining the objective functions for cross-modal similarity $\mathcal{L}_{S}$, inter-concept decoupling $\mathcal{L}_{D}$ and intra-concept alignment $\mathcal{L}_{A}$ mentioned above, we get the total training loss $\mathcal{L}=\mathcal{L}_{S}+\alpha \mathcal{L}_{D}+\beta \mathcal{L}_{A}$, where $\alpha$ and $\beta$ are the trade-off hyper-parameters.

\begin{figure*}[thb]
    \centering
    \includegraphics[width=1\linewidth]{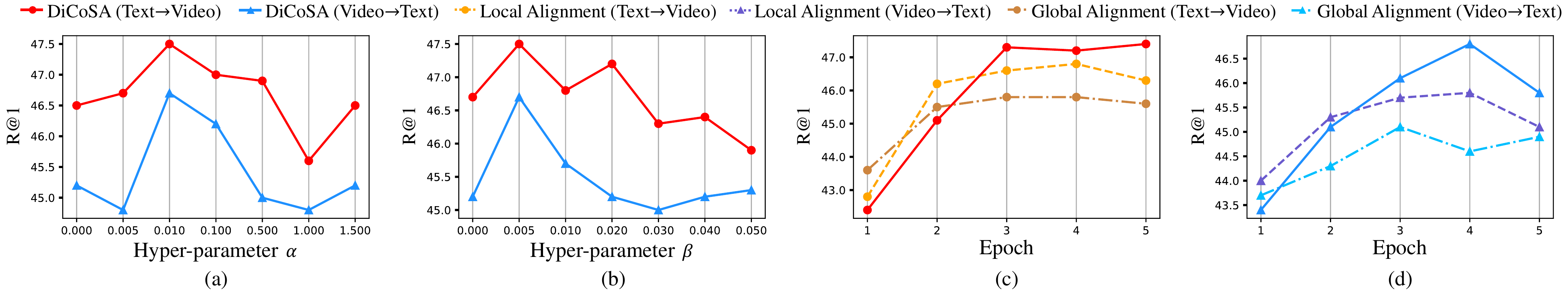}
    \vspace{-2.0em}
    \caption{\textbf{Ablation study on the MSRVTT dataset.} Effect of (a) the trade-off hyper-parameter $\alpha$; (b) the trade-off hyper-parameter $\beta$. The comparison with the global alignment and the local alignment (c) on the text-to-video task; (d) on the video-to-text task.}
    \label{fig:experiment0}
\end{figure*}

\begin{table*}
\begin{minipage}[c]{0.68\textwidth}
\centering
\resizebox{1.0\linewidth}{!}{
\footnotesize
\begin{tabular}{l|ccc|ccc}
\toprule[1.5pt]
\multirow{2}{*}{\textbf{Method}}  & \multicolumn{3}{c|}{\textbf{Text-\textgreater{}Video}}        & \multicolumn{3}{c}{\textbf{Video-\textgreater{}Text}} \\ 
\cmidrule(rl){2-4}\cmidrule(rl){5-7}
 & R@1$\uparrow$ & R@5$\uparrow$ & MnR$\downarrow$
& R@1$\uparrow$ & R@5$\uparrow$  & MnR$\downarrow$    \\ \midrule
Baseline & 45.9 & 72.0 & 14.1 & 45.2 & 73.5 & 9.8 \\
+ Latent Factors  & 46.2 & 73.9 & 14.0 & 46.0 & 73.6 & 10.2 \\
+ Inter-Concept Decoupling $\mathcal{L}_{D}$  & 46.5 & 73.4 & \textbf{13.2} & 46.1 & 74.1 & 9.9 \\
+ Intra-Concept Alignment $\mathcal{L}_{A}$ & 47.1 & 73.6 & 13.3 & 46.1 & 74.0 & 10.0\\
\rowcolor{aliceblue!60} + Adaptive Pooling  & \textbf{47.5} & \textbf{74.7} & \textbf{13.2} & \textbf{46.7} & \textbf{75.2} & \textbf{8.9} \\ \bottomrule[1.5pt]
\end{tabular}
}
\vspace{-.5em}
\caption{\textbf{Ablation study for the architecture design  on the MSR-VTT dataset.} ``Baseline'' denotes the global alignment. ``$\uparrow$'' denotes that higher is better. ``$\downarrow$'' denotes that lower is better.}
\label{Experiments1}
\vspace{1.0em}
\end{minipage}
\hfill
\begin{minipage}[c]{0.3\textwidth}
\centering
\resizebox{1.0\linewidth}{!}{
\footnotesize
\begin{tabular}{l|cccc}
\toprule[1.5pt]
\multirow{2}{*}{\textbf{Method}}  & \multicolumn{3}{c}{\textbf{Text-\textgreater{}Video}} \\ 
\cmidrule(rl){2-4}
 & R@1$\uparrow$ & R@5$\uparrow$ & MnR$\downarrow$ \\ \midrule
K=2 & 46.2 & 72.3 & 13.2\\
K=4  & 46.7 & 73.0 & 13.0 \\
\rowcolor{aliceblue!60} K=8  & \textbf{47.5} & \textbf{74.7} & 13.2 \\
K=16  & 44.6 & 72.0 & \textbf{12.9} \\
K=32  & 43.2 & 71.5 & 14.5 \\ \bottomrule[1.5pt]
\end{tabular}
}
\vspace{-.5em}
\caption{\textbf{Ablation study for the number of concepts $K$ on the MSR-VTT dataset.}}
\label{Experiments2}
\vspace{1.0em}
\end{minipage}
\centering
\resizebox{1.0\linewidth}{!}{
\begin{tabular}{l|cc|ccccc|ccccc}
\toprule[1.5pt]
\multirow{2}{*}{\textbf{Method}} &\multirow{2}{*}{\textbf{Complexity}}  & \multirow{2}{*}{\textbf{Time (ms)}} & \multicolumn{5}{c|}{\textbf{Text-\textgreater{}Video}}        & \multicolumn{5}{c}{\textbf{Video-\textgreater{}Text}} \\ 
\cmidrule(rl){4-8}\cmidrule(rl){9-13}
 & & & R@1$\uparrow$ & R@5$\uparrow$ & R@10$\uparrow$ & MdR$\downarrow$ & MnR$\downarrow$
& R@1$\uparrow$ & R@5$\uparrow$ & R@10$\uparrow$ & MdR$\downarrow$ & MnR$\downarrow$  \\ \midrule
Global Alignment & \bm{$\mathcal{O}(D)$} & \textbf{806} & 45.9 & 72.0 & 81.7 & \textbf{2.0} & 14.1 & 45.2 & 73.5 & 82.6 & \textbf{2.0} & 9.8 \\
Local Alignment & $\mathcal{O}(N_tN_vD)$ & 1158 & 46.8 & 72.6 & 82.6 & \textbf{2.0} & 13.4 & 46.4 & 72.2 & 82.3 & \textbf{2.0} & 13.4\\
\midrule
\rowcolor{aliceblue!60} \textbf{DiCoSA (Ours)}  & \bm{$\mathcal{O}(D)$} & 978 & \Frst{47.5} & \Frst{74.7} & \Frst{83.8} & \Frst{2.0} & \Frst{13.2} & \Frst{46.7} & \Frst{75.2} & \Frst{84.3} & \Frst{2.0} & \Frst{8.9} \\ \bottomrule[1.5pt]
\end{tabular}
}
\vspace{-.5em}
\caption{\textbf{The comparison with the global alignment and the local alignment on the MSR-VTT dataset.} ``$\uparrow$'' denotes that higher is better. ``$\downarrow$'' denotes that lower is better. We report the average inference time for processing the test set (1k videos and 1k text queries) using two Tesla V100 GPUs. Here, $D$, $N_v$ and $N_t$ denote the feature dimension, the frame length and the text length, respectively.}
\label{Experiments3}
\end{table*}

\subsection{Intuitive Analysis}
Current methods mainly perform global alignment or local alignment. Now we explain other advantages of the proposed DiCoSA besides disentangled representation, mainly in three aspects. \textbf{(i) Humanlike set-to-set matching.} Humans perceive the world by conceptualizing high-dimensional inputs from multiple modalities such as vision and language. Through conceptualization, humans integrate things into conceptual networks and make inferences based on them. Our method simulates the human process of conceptualizing things and reasoning text and video with concepts to achieve set-to-set cross-modal matching. \textbf{(ii) Avoiding curse of dimension.} As the dimension of representation space increases, the model becomes more and more difficult to optimize, which is known as the curse of dimension~\cite{kuo2005lifting}. Therefore, we choose to optimize the decoupled subspace ($\mathbb{R}^{\frac{D}{K}}$) at lower dimensions rather than directly learn the representation space ($\mathbb{R}^D$), which we will discuss in experiments (see Figure~{\color{red}\ref{fig:experiment0}} (c) and (d)). \textbf{(iii) Efficiency.} Our method combines efficiency and granularity well. We calculate inference time on the MSRVTT dataset, which we will discuss in experiments (see Table~{\color{red}\ref{Experiments3}}).

\section{Experiments}
\subsection{Experiment Setup}
\textbf{Datasets.} \textbf{MSR-VTT}~\cite{xu2016msr} contains 10,000 YouTube videos, each with 20 text descriptions. We follow the 1k-A split~\cite{liu2019use} with 9,000 videos for training and 1,000 for testing. \textbf{LSMDC}~\cite{rohrbach2015a} contains 118,081 video clips from 202 movies. We follow the split of \cite{gabeur2020multi} with 1,000 videos for testing. \textbf{MSVD}~\cite{chen2011collecting} contains 1,970 videos. We follow the official split of 1,200 and 670 as the train and test set, respectively. \textbf{ActivityNet Caption}~\cite{krishna2017dense} contains 20K YouTube videos. We report results on the ``val1'' split of 10,009 and 4,917 as the train and test set. \textbf{DiDeMo}~\cite{anne2017localizing} contains 10k videos annotated 40k text descriptions. We follow the training and evaluation protocol in \cite{luo2021clip4clip}.

\noindent \textbf{Metrics.} We choose Recall at rank L (R@L, higher is better), Median Rank (MdR, lower is better) and mean rank (MnR, lower is better) to evaluate the performance.

\noindent \textbf{Implementation Details.} Following previous works~\cite{luo2021clip4clip,liu2022ts2,jin2022expectation}, we utilize the CLIP (ViT-B/32)~\cite{radford2021learning} as the pre-trained model. The dimension of the feature is 512. The temporal transformer~\cite{vaswani2017attention,li2022locality} is composed of 4-layer blocks, each including 8 heads and 512 hidden channels. The temporal position embedding and parameters are initialized from the CLIP’s text encoder. We use the Adam optimizer~\cite{kingma2014adam} with a linear warmup. The initial learning rate is 1e-7 for the text encoder and video encoder and 1e-3 for other modules. If not otherwise specified, we set $\tau^{'} = 0.01$, $K = 8$, $\alpha = 0.01$, $\beta=0.005$. The network is optimized with the batch size of 128 in 5 epochs. During the inferring phase, we assume that only the candidate set is known in advance. We follow inferring schedules from~\cite{bogolin2022cross}. More details are in the Appendix.

\begin{figure*}[thb]
    \centering
    \includegraphics[width=1.0\linewidth]{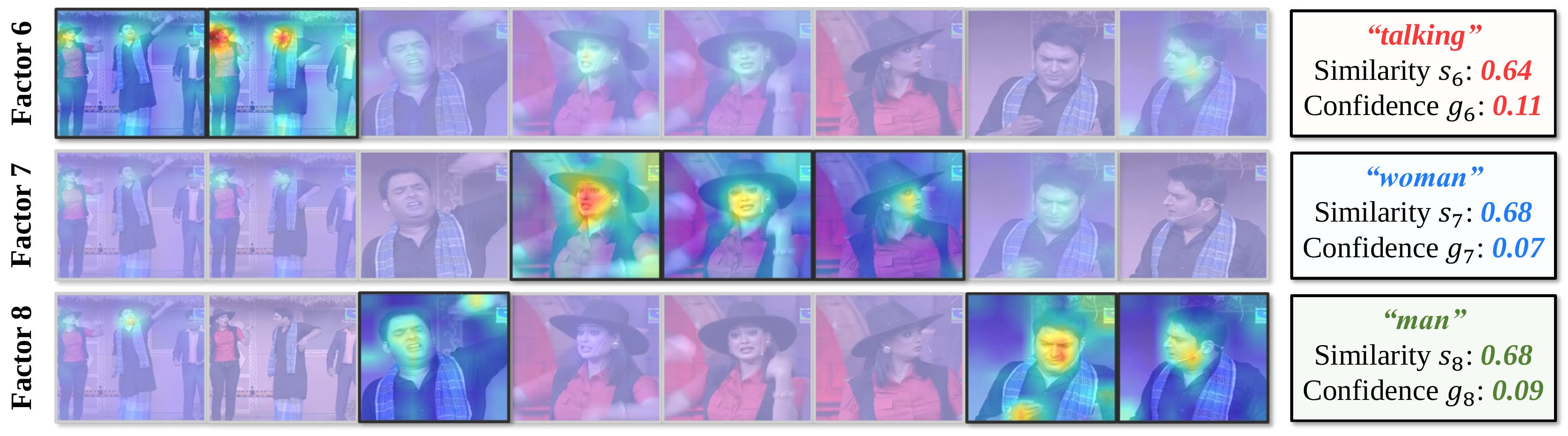} 
    \vspace{-1.3em}
    \caption{\textbf{Attention visualization of latent factors.} We take Video0070 in the MSR-VTT dataset as an example.  Each latent factor learned by our method individually focuses on a specific concept. Specifically, Factor 6 focuses on \textcolor{cred}{\textbf{``talking''}}; Factor 7 notices \textcolor{cblue}{\textbf{``woman''}}; and Factor 7 corresponds to \textcolor{cgreen}{\textbf{``man''}}. $s_k$ is the similarity of the $k_{th}$ factor pair. $g_k$ denotes the estimated confidence of the $k_{th}$ factor pair.}%The implementation details of the visualization are in the supplementary material.}
    \label{fig:experiment1}
    \vspace{-.5em}
\end{figure*}

\subsection{Comparisons with State-of-the-art Methods}
We compare the proposed DiCoSA with other methods on five benchmarks. In Table~{\color{red}\ref{MSRVTT results}}, we show the results of our method on the MSR-VTT dataset. Our model outperforms the recently proposed SOTA methods on both text-to-video retrieval and video-to-text retrieval tasks.  Table~{\color{red}\ref{Experiments0}} shows text-to-video retrieval results on the LSMDC, MSVD, ActivityNet and DiDeMo datasets. The results on all five datasets demonstrate that our method is capable of dealing with both short and long videos. DiCoSA achieves consistent improvements across different datasets, which demonstrates the effectiveness and generalization ability of our method. 

\subsection{Ablation Study}
\noindent \textbf{Architecture Design.} 
To illustrate the importance of each part of our method, we conduct ablation experiments on the MSR-VTT dataset. From Table~{\color{red}\ref{Experiments1}}, we can draw the following observations: (i)~The model using latent factors achieves comparable or better performance than the baseline on the two retrieval tasks. We consider that it is because the latent factor dimension is small, thus alleviating the curse of dimension. (ii)~Compared with the optimization of latent factors only from the inter-concept perspective, the improvement of the optimization from both inter-concept and intra-concept perspectives is more significant. We consider that it is because intra-concept alignment avoids the interference of mismatched cross-modal concepts in the learning of other concepts. (iii)~``Adaptive pooling'' can locate the mismatched cross-modal concepts and improve the respective discriminative power for set-to-set matching. Our full model achieves the best performance. This demonstrates that the four parts are beneficial for aligning visual contents and textual semantics.

\noindent \textbf{Effect of the Number of Concepts.} 
The concept size $K$ controls the number of latent factors $\bm{E}=[e_1,e_2,...,e_K]$, where $e_k\in \mathbb{R}^{\frac{D}{K}}$. We start with a small size and increase it to large ones. In Table~{\color{red}\ref{Experiments2}}, overall performance improves and then decreases. On the one hand, we find that fewer concepts limit the ability to leverage fine-grained information. On the other hand, a larger number of concepts reduces the dimension of each latent factor $e_k\in \mathbb{R}^{\frac{D}{K}}$, which limits the discriminability of the factors. We set the concept size $K = 8$ to achieve the best performance in practice.

\noindent \textbf{Parameter Sensitivity.} 
The parameter $\alpha$ indicates the importance of $\mathcal{L}_{D}$. We evaluate the scale range setting $\alpha \in [0.0, 1.5]$ as shown in Figure~{\color{red}\ref{fig:experiment0}} (a). We find that R@1 is improved from 46.7\% to 47.5\% when $\alpha = 0.005$ and saturated with $\alpha = 0.01$ for text-to-video retrieval. As a result, we adopt $\alpha = 0.01$ to achieve the best performance. In Figure~{\color{red}\ref{fig:experiment0}} (b), we show the influence of the hyper-parameter $\beta$. We evaluate the scale range setting $\beta \in [0.0, 0.05]$. We find that the model achieves the best performance at $\beta=0.005$, so we set $\beta=0.005$ as the default in practice.

\noindent \textbf{Comparisons to Other Baseline Methods.} 
We further compare our method with other baseline methods in Table~{\color{red}\ref{Experiments3}}. Since the dimension of latent factor ($\mathbb{R}^{\frac{D}{K}}$) is lower than the original feature dimension ($\mathbb{R}^{D}$), our method introduces negligible computational overhead. Moreover, our method brings remarkable improvements by disentangled conceptualization and set-to-set alignment. We also show the training process of our method and baseline methods in Figure~{\color{red}\ref{fig:experiment0}} (c) and (d). Because we optimize the decoupled subspace at a lower dimension ($\mathbb{R}^{\frac{D}{K}}$), our method has the most efficient performance improvement. We observe that the performance of our method in early epochs is lower than other methods because our disentangled alignment is far removed from the CLIP pre-training objective and therefore cannot directly transfer the knowledge of CLIP in early epochs.

\subsection{Qualitative Analysis}
To better understand the proposed method, we show the visualization of latent factors in Figure~{\color{red}\ref{fig:experiment1}}. In our method, each latent factor individually focuses on a specific detail. Specifically, Factor 6 focuses on ``talking''; Factor 7 notices ``woman''; and Factor 7 corresponds to ``man''. Our latent factors are explainable to some extent. Therefore, the proposed method can be used as a tool for visualizing the cross-modal interaction and help us understand the existing retrieval model. Interestingly, we find that the factor pair corresponding to action~(``talking'') has higher confidence than those corresponding to entities~(``woman'' and ``man''). This result illustrates that the model tends to judge cross-modal similarity by actions rather than entities. The implementation details of the visualization are in the supplementary material.

\section{Conclusion}
In this paper, we propose the Disentangled Conceptualization and Set-to-set Alignment~(DiCoSA) for text-video retrieval. DiCoSA simulates the conceptualizing and reasoning process of human beings. Moreover, it is superior at computation efficiency and granularity, ensuring fine-grained interactions with local alignment using a similar computational complexity as global alignment. Experimental results on five text-video retrieval benchmark datasets show the advantages of the proposed method. Further, it is worth noting that our interpretable latent factors can reflect the cross-modal interaction. In the future, we hope that the disentangled features learned by our method could also be applied to other cross-modal tasks.

\noindent \textbf{Acknowledgements.} This work was supported in part by the National Key R\&D Program of China (No. 2022ZD0118201), Natural Science Foundation of China (No. 61972217, 32071459, 62176249, 62006133, 62271465), and the Natural Science Foundation of Guangdong Province in China (No. 2019B1515120049).

\bibliographystyle{named}
\bibliography{bib}

\clearpage
\appendix
\renewcommand{\thetable}{\Alph{table}}
\renewcommand{\theequation}{\Alph{equation}}
\renewcommand{\thefigure}{\Alph{figure}}
\setcounter{table}{0}
\setcounter{section}{0}
\setcounter{figure}{0}
\setcounter{equation}{0}

\section{Datasets}
We compare the proposed \underline{Di}sentangled \underline{Co}nceptualization and \underline{S}et-to-set \underline{A}lignment~(DiCoSA) with other methods on five benchmark text-video retrieval datasets, including MSR-VTT, LSMDC, MSVD, ActivityNet, and DiDeMo.

\begin{itemize}
\item \textbf{MSR-VTT}~\cite{xu2016msr} contains 10,000 YouTube videos, each with 20 text descriptions. Videos in the MSR-VTT dataset are medium in duration, lasting about 10 to 30 seconds. We follow the 1k-A split~\cite{liu2019use} with 9,000 videos for training and 1,000 for testing. 

\item \textbf{LSMDC}~\cite{rohrbach2015a} contains 118,081 video clips from 202 movies. The duration of videos in the LSMDC dataset is short. We follow the split of \cite{gabeur2020multi} with 1,000 videos for testing. 

\item \textbf{MSVD}~\cite{chen2011collecting} contains 1,970 videos. Each video has approximately 40 associated text description. Videos in the MSVD dataset are short in duration, lasting about 10 to 25 seconds. We follow the official split of 1,200 and 670 as the train and test set, respectively. 

\item \textbf{ActivityNet Caption}~\cite{krishna2017dense} contains 20K YouTube videos. Videos in the ActivityNet dataset are long in duration, lasting about 180 seconds.  Following previous works~\cite{luo2021clip4clip,jin2022expectation}, we evaluate video-paragraph retrieval, where all sentence descriptions for a video are concatenated into a single query. We report results on the ``val1'' split of 10,009 and 4,917 as the train and test set. 

\item \textbf{DiDeMo}~\cite{anne2017localizing} contains 10,464 videos annotated 40,543 text descriptions. The duration of videos in the DiDeMo dataset is short. Following previous works~\cite{luo2021clip4clip}, we evaluate video-paragraph retrieval, where all sentence descriptions for a video are concatenated into a single query. We follow the training and evaluation protocol in \cite{luo2021clip4clip}.
\end{itemize}

\section{Implementation Details}
Following previous works~\cite{luo2021clip4clip,liu2022ts2,jin2022expectation}, we utilize the CLIP (ViT-B/32)~\cite{radford2021learning} as the pre-trained model. The dimension of the feature is 512. The temporal transformer is composed of 4-layer blocks, each including 8 heads and 512 hidden channels. The temporal position embedding and parameters are initialized from the CLIP’s text encoder. We use the Adam optimizer~\cite{kingma2014adam} with a linear warmup. The initial learning rate is 1e-7 for the text encoder and video encoder and 1e-3 for other modules. If not otherwise specified, we set $\tau = 3$, $\tau^{'} = 0.01$, $K = 8$, $\alpha = 0.01$, $\beta=0.005$. For short video retrieval datasets, i.e., MSR-VTT, LSMDC and MSVD, the word length is 32 and the frame length is 12. For long video retrieval datasets, i.e., ActivityNet Caption and DiDeMo, we set the word length to 64 and the frame length is 64. The network is optimized with the batch size of 128 in 5 epochs. We use 8 V100 GPUs to train the network. During the inferring phase, we assume that only the candidate set is known in advance. We follow inferring schedules from~\cite{bogolin2022cross}. We have attached the code into the supplementary material and promise to release it upon publication.

\section{Proof of Lemma 1}
To capture disentangled representation, we optimize latent factors from both inter-concept and intra-concept perspectives. To this end, we consider the mutual information between any two latent factors $e_i$ and $e_j$:
\begin{equation}
I(e_i;e_j)=\mathbb{E} \left[ p(e_i,e_j)\log\frac{p(e_i,e_j)}{p(e_i)p(e_j)}\right].
\end{equation}
Obviously, the mutual information is hard to measure directly. To this end, we convert the problem to minimizing $\frac{p(e_i,e_j)}{p(e_i)p(e_j)}$. Concretely, given latent factors $e_i\in \mathbb{R}^{\frac{D}{K}}$ and $e_j\in \mathbb{R}^{\frac{D}{K}}$, we first normalize them by the following formula:
\begin{equation}
z_i=\frac{e_i-\mathbb{E}\left[e_i\right]}{\sqrt{\textrm{Var}\left[e_i\right] }},\quad z_j=\frac{e_j-\mathbb{E}\left[e_j\right]}{\sqrt{\textrm{Var}\left[e_j \right] }},
\end{equation}
where $z_i, z_j$ have the same mean and standard deviation. In this manner, we scale the latent factors to the standard scale. Then, we calculate the covariance of $z_i$ and $z_j$ as follows,
\begin{equation}
C_{i,j}=\mathbb{E} \left[ (z_i)^{\top}z_j \right],
\end{equation}
where $z_i$ and $z_j$ are the normalized features of $e_i$ and $e_j$, respectively. We start our analysis with the joint probability $p(e_i,e_j)$. Specifically, $p(e_i,e_j)$ can be expanded as follows,
\begin{equation}
\begin{aligned}
p(e_i,e_j)&=\frac{p(e_i|e_j)\footnotesize{\prod_{l\neq i}}p(e_l)}{\footnotesize{\sum_{k=1}^{K}}p(e_k|e_j) \prod_{l\neq k}p(e_l)}\\
&=\frac{\frac{p(e_i|e_j)}{p(e_i)}}{\footnotesize{\sum_{k=1}^{K}}\frac{p(e_k|e_j)}{p(e_k)}}\\
&=\frac{\frac{p(e_i,e_j)}{p(e_i)p(e_j)}}{\footnotesize{\sum_{k=1}^{K}}\frac{p(e_k,e_j)}{p(e_k)p(e_j)}}.
\label{p1}
\end{aligned}
\end{equation}
Meanwhile, the meaning of $p(e_i,e_j)$ is the probability that $e_i$ matches $e_j$. So it can be expressed as:
\begin{equation}
\begin{aligned}
p(e_i,e_j)&=\frac{\mathbb{E} \left[ (z_i)^{\top}z_j \right]}{\sum_{k=1}^{K}\mathbb{E} \left[ (z_k)^{\top}z_j \right]}\\
&=\frac{C_{i,j}}{\sum_{k=1}^{K}C_{k,j}}.
\label{p2}
\end{aligned}
\end{equation}
As shown in Equation~{\color{red}\ref{p1}} and Equation~{\color{red}\ref{p2}}, maximizing (minimizing) $\frac{p(e_i,e_j)}{p(e_i)p(e_j)}$ is equivalent to maximizing (minimizing) $C_{i,j}$, i.e., $C_{i,j} \propto \frac{p(e_i,e_j)}{p(e_i)p(e_j)}$. Based on the above analysis, we can minimize $I(e_i;e_j)$ by the following loss,
\begin{equation}
\mathcal{L}_{D}=\sum_i \sum_{j\neq i} (C_{i,j})^2.
\end{equation}
Similarly, we can maximize $I(e_i;e_i)$ by the following loss,
\begin{equation}
\mathcal{L}_{A}=\sum_i (1-C_{i,i})^2.
\end{equation}

\begin{figure*}[thb]
    \centering
    \includegraphics[width=1.0\linewidth]{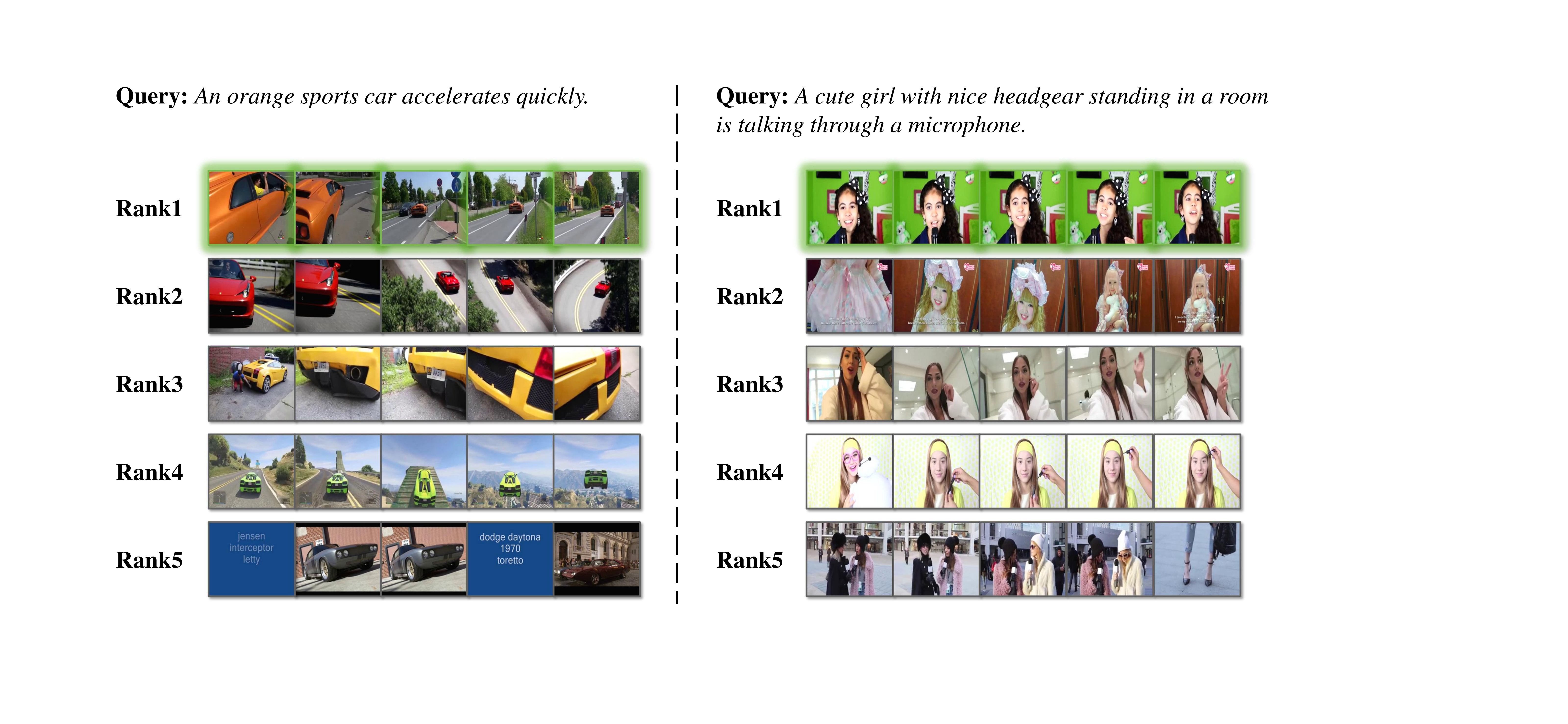} 
    \caption{\textbf{Visualization of the text-to-video results.} Only the correct videos are highlighted in green.}
    \label{fig:appendix1}
\end{figure*}

\section{Details of the Attention Map Visualization}
We visualize the attention map in two steps: (i) At every Transformer block, we get an attention matrix that defines how much attention is going to flow from the token in the previous layer to the token in the next layer. We multiply the matrices between every two layers to get the total attention flow between them. (ii) For one factor, we weigh all the attention by this factor gradient and then take the average among the attention heads.

\section{Visualization of the Text-to-Video Results} We show two examples of the videos retrieved by our method in Figure~{\color{red}{\ref{fig:appendix1}}}. As shown in Figure~{\color{red}{\ref{fig:appendix1}}}, our DiCoSA successfully retrieves the ground-truth video.
\end{document}